\documentclass[sigconf]{acmart}
\usepackage{amsmath,amssymb,amsfonts,bm,amsthm}
\usepackage[linesnumbered,ruled]{algorithm2e}
\usepackage{graphicx}
\usepackage[tight,normal]{subfigure}
\usepackage{multirow,makecell,url,footmisc}
\usepackage{enumerate}
\usepackage{lineno}
\usepackage{textcomp}
\usepackage{booktabs} 
\usepackage{arydshln}
\usepackage{url}
\usepackage{float}

\renewcommand\footnotetextcopyrightpermission[1]{}
\copyrightyear{2018} 
\acmYear{2018} 
\setcopyright{acmcopyright}
\acmConference[KDD '18]{The 24th ACM SIGKDD International Conference on Knowledge Discovery & Data Mining}{August 19--23, 2018}{London, United Kingdom}
\acmBooktitle{KDD '18: The 24th ACM SIGKDD International Conference on Knowledge Discovery & Data Mining, August 19--23, 2018, London, United Kingdom}
\acmPrice{15.00}
\acmDOI{10.1145/3219819.3219826}
\acmISBN{978-1-4503-5552-0/18/08}

\begin{document}
\title{Learning Tree-based Deep Model for Recommender Systems}

\author{Han Zhu, Xiang Li, Pengye Zhang, Guozheng Li, Jie He, Han Li, Kun Gai}
\affiliation{%
\institution{Alibaba Group}
}
\email{{zhuhan.zh, yushi.lx, pengye.zpy, guozheng.lgz, jay.hj, lihan.lh, jingshi.gk}@alibaba-inc.com}

\begin{abstract}
    
    Model-based methods for recommender systems have been studied extensively in recent years.
    In systems with large corpus, however, 
    the calculation cost for the learnt model to predict all user-item preferences is tremendous,
    which makes full corpus retrieval extremely difficult.
    To overcome the calculation barriers, models such as matrix factorization
    resort to inner product form (i.e., model user-item preference as the inner product of user, item latent factors) and
    indexes to facilitate efficient approximate k-nearest neighbor searches.
    However, it still remains challenging to incorporate more expressive interaction forms between user and item features, 
    e.g., interactions through deep neural networks, because of the calculation cost.

    In this paper, we focus on the problem of introducing arbitrary advanced models to recommender systems with large corpus.
    We propose a novel tree-based method which can provide logarithmic complexity w.r.t. corpus size even with more expressive models such as deep neural networks.
    Our main idea is to predict user interests from coarse to fine by traversing tree nodes in a top-down fashion and making decisions for each user-node pair.
    We also show that the tree structure can be jointly learnt towards better compatibility with users' interest distribution and hence facilitate both training and prediction.
    Experimental evaluations with two large-scale real-world datasets show that the proposed method significantly outperforms traditional methods.
    Online A/B test results in Taobao display advertising platform also demonstrate the effectiveness of the proposed method in production environments.

\end{abstract}



\maketitle

\section{Introduction}\label{section:Introduction}


Recommendation has been widely used by various kinds of content providers. 
Personalized recommendation method, based on the intuition that users' interests can be inferred from their historical behaviors 
or other users with similar preference, has been proven to be effective in YouTube \cite{Covington2016Deep} and Amazon \cite{linden2003amazon}.


Designing such a recommendation model to predict the best candidate set from the entire corpus for each user has many challenges. 
In systems with enormous corpus, some well-performed recommendation algorithms may fail to predict from the entire corpus. 
The linear prediction complexity w.r.t. the corpus size is unacceptable.
Deploying such large-scale recommender system requires the amount of calculation to predict for each single user be limited.
And besides preciseness, the novelty of recommended items should also be responsible for user experience.
Results that only contain homogeneous items with user's historical behaviors are not expected.


To reduce the amount of calculation and handle enormous corpus, memory-based collaborative filtering methods are widely deployed in industry \cite{linden2003amazon}.
As a representative method in collaborative filtering family, item-based collaborative filtering \cite{Sarwar2001Item} can recommend 
from very large corpus with relatively much fewer computations, 
depending on the pre-calculated similarity between item pairs and using user's historical behaviors as triggers to recall those most similar items.
However, there exists restriction on the scope of candidate set, i.e., 
not all items but only items similar to the triggers can be ultimately recommended. 
This intuition prevents the recommender system from jumping out of historical behavior to explore potential user interests,
which limits the accuracy of recalled results.
And in practice the recommendation novelty is also criticized.
Another way to reduce calculation is making coarse-grained recommendation. 
For example, the system recommends a small number of item categories for users and picks out all corresponding items, 
with a following ranking stage. However, for large corpus, the calculation problem is still not solved. 
If the category number is large, the category recommendation itself also meets the calculation barrier. 
If not, some categories will inevitably include too many items, making the following ranking calculation impracticable. 
Besides, the used categories are usually not designed for recommendation problem, which can seriously harm the recommendation accuracy.


In the literatures of recommender systems, model-based methods are an active topic.
Models such as matrix factorization (MF) \cite{Koren2009Matrix, Salakhutdinov2007Probabilistic} try to decompose pairwise user-item preferences (e.g., ratings) into user and item factors, 
then recommend to each user its most preferred items. 
Factorization machine (FM) \cite{Rendle2010Factorization} further proposes a unified model 
that can mimic different factorization models with any kind of input data.
In some real-world scenarios that have no explicit preference but only implicit user feedback (e.g., user behaviors like clicks or purchases), Bayesian personalized ranking \cite{rendle2009bpr} gives a solution that 
formulates the preference in triplets with partial order, and applies it to MF models. 
In industry, YouTube uses deep neural network \cite{Covington2016Deep} to learn both user and item's embeddings,
where two kinds of embeddings are generated from their corresponding features separately.
In all the above kinds of methods, the preference of user-item pair can be formulated as the inner product of user and item's vector representations.
The prediction stage thus is equivalent to retrieve user vector's nearest neighbors in inner product space.
For vector search problem, 
indices like hashing or quantization \cite{JDH17} for approximate k-nearest neighbor (kNN) search can ensure the efficiency of retrieval.

However, the inner product interaction form between user and item's vector representations severely limits model's capability.
There exist many other kinds of more expressive interaction forms,
for example, cross-product features between user's historical behaviors and candidate items are widely used in click-through rate prediction \cite{cheng2016wide}.
Recent work \cite{he2017neural} proposes a neural collaborative filtering method, 
where a neural network instead of inner product is used to model the interaction between user and item's vector representations.
The work's experimental results prove that a multi-layer feed-forward neural network performs better than the fixed inner product manner.
Deep interest network \cite{Zhou2017Deep} points out that user interests are diverse,
and an attention like network structure can generate varying user vectors according to different candidate items.
Beyond the above works, other methods like product neural network \cite{Qu2016Product} 
have also proven the effectiveness of advanced neural networks.
However, as these kinds of models can not be regulated to inner product form between user and item vectors to utilize efficient approximate kNN search, 
they can not be used to recall candidates in large-scale recommender systems.
How to overcome the calculation barrier to make arbitrary advanced neural networks feasible in large-scale recommendation is a problem.


To address the challenges above, we propose a novel tree-based deep recommendation model (TDM) in this paper.
Tree and tree-based methods are researched in multiclass classification problem \cite{Bengio2010Label, Choromanska2015Logarithmic, Beygelzimer2007Multiclass, prabhu2014fastxml, weston2013label, agrawal2013multi, jain2016extreme},
where tree is usually used to partition the sample or label space to reduce calculation cost.
However, researchers seldom set foot in the context of recommender systems using tree structure as an index for retrieval. 
Actually, hierarchical structure of information ubiquitously exists in many domains.
For example, in E-commerce scenario, iPhone is the fine-grained item while smartphone is the coarse-grained concept to which iPhone belongs.
The proposed TDM method leverages this hierarchy of information and turns recommendation problem into a series of hierarchical classification problems.
By solving the problem from easy to difficult, TDM can improve both accuracy and efficiency.
The main contributions of our paper are summarized as follows:

\begin{itemize}
    \item To our best knowledge, TDM is the first method that makes arbitrary advanced models possible in generating recommendations from large corpus.
    Benefiting from hierarchical tree search, TDM achieves logarithmic amount of calculation w.r.t. corpus size when making prediction. 

    \item TDM can help find novel but effective recommendation results more precisely,
    because the entire corpus is explored and more effective deep models also can help find potential interests.

    \item Besides more advanced models, TDM also promotes recommendation accuracy by hierarchical search, 
    which divides a large problem into smaller ones and solves them successively from easy to difficult.

    \item As a kind of index, the tree structure can also be learnt towards optimal hierarchy of items and concepts for more effective retrieval, 
    which in turn facilitates the model training. 
    We employ a tree learning method that allows joint training of neural network and the tree structure.

    \item We conduct extensive experiments on two large-scale real-world datasets, 
    which show that TDM outperforms existing methods significantly.
\end{itemize}


It's worth mentioning that tree-based approach is also researched in language model work hierarchical softmax \cite{Morin2005Hierarchical}, 
but it's different from the proposed TDM not only in motivation but also in formulation.
In next-word prediction problem, conventional softmax has to calculate the normalization term to get any single word's probability, which is very time-consuming.
Hierarchical softmax uses tree structure, and next-word's probability is converted to the product of node probabilities along the tree path.
Such formulation reduces the computation complexity of next-word's probability to logarithmic magnitude w.r.t. the corpus size.
However, in recommendation problem, the goal is to search the entire corpus for those most preferred items, which is a retrieval problem.
In hierarchical softmax tree, the optimum of parent nodes can not guarantee that the optimal low level nodes are in their descendants, 
and all items still need to be traversed to find the optimal one. 
Thus, it's not suitable for such a retrieval problem.
To address the retrieval problem, we propose a max-heap like tree formulation and introduce deep neural networks to model the tree, 
which forms an efficient method for large-scale recommendation.
The following sections will show its difference in formulation and its superiority in performance.
In addition, hierarchical softmax adopts a single hidden layer network for a specific natural language processing problem, 
while the proposed TDM method is practicable to engage any neural network structures.

The proposed tree-based model is a universal solution for all kinds of online content providers.
The remainder of this paper is organized as follows: In Section~\ref{section:Architecture}, we'll introduce the system architecture
of Taobao display advertising to show the position of the proposed method. Section~\ref{section:Mechanism} will give a detailed introduction and formalization of 
the proposed tree-based deep model. And the following Section~\ref{section:Serving} will describe how the tree-based model serves online.
Experimental results on large-scale benchmark dataset and Taobao advertising dataset are shown in Section~\ref{section:Experiment}. At last, 
Section~\ref{section:Conclusion} gives our work a conclusion.

\section{System Architecture} \label{section:Architecture}
\begin{figure}[tb]
    \centering
    \includegraphics[width=1.0\columnwidth]{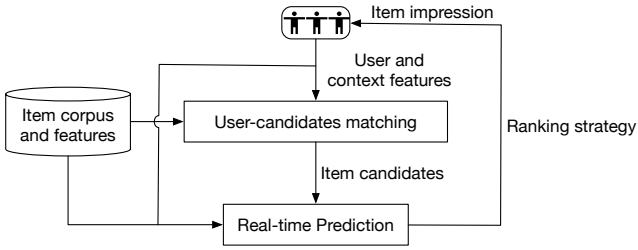}
    \caption{The system architecture of Taobao display advertising recommender system.}
    \label{fig:Architechture}
\end{figure}

In this section, we introduce the architecture of Taobao display advertising recommender system as Figure~\ref{fig:Architechture}.
After receiving page view request from a user, the system uses user features, context features and item features as
input to generate a relatively much smaller set (usually hundreds) of candidate items from the entire corpus (hundreds of millions) in the matching server. 
The tree-based recommendation 
model takes effort in this stage and shrinks the size of candidate set by several orders of magnitude. 

With hundreds of candidate items, the real-time prediction server uses more expressive but also more time consuming models \cite{Zhou2017Deep, ge2017image}
to predict indicators like click-through rate or conversion rate. And after ranking by strategy \cite{OCPC, jin2018real}, several items are ultimately impressed to user.

As aforementioned, the proposed recommendation model aims to construct a candidate set with hundreds of items. This stage is essential and also difficult.
Whether the user is interested in the generated candidates gives an upper bound of the impression quality.
How to draw candidates from the entire corpus weighing efficiency and effectiveness is a problem.

\section{Tree-based Deep Model} \label{section:Mechanism}
In this part, we first introduce the tree structure used in our tree-based model to give an overall conception.
Secondly, we introduce hierarchical softmax \cite{Morin2005Hierarchical} to show why its formulation is not suitable for recommendation.
After that, we give a novel max-heap like tree formulation and show how to train the tree-based model.
Then, the deep neural network architecture is introduced.
At last, we show how to construct and learn the tree used in the tree-based model.

\subsection{Tree for Recommendation} \label{section:TreeForRecommendation}
A recommendation tree consists of a set of nodes $N$, where 
$N = \{n_1, n_2, \cdots n_{|N|}\}$ represents $|N|$ individual non-leaf or leaf nodes.
Each node in $N$ except the root node has one parent and an arbitrary number of children. 
Specifically, each item $c_i$ in the corpus $C$ corresponds to one and only one \emph{leaf node} in the tree,
and those non-leaf nodes are coarse-grained concepts.
Without loss of generality, we suppose that node $n_1$ is always the root node.
An example tree is illustrated in the right bottom corner of Figure~\ref{fig:Model}, in which 
each circle represents a node and the number of node is its index in tree. The tree has 
$8$ leaf nodes in total, each of which corresponds to an item in the corpus. It's worth mentioning that though the given
example is a complete binary tree, we don't impose complete and binary as restrictions on the type of the tree in our model.

\subsection{Related Work} \label{section:RelatedWork}
With the tree structure, we firstly introduce the related work hierarchical softmax to help understand its difference with our TDM.
In hierarchical softmax, each leaf node $n$ in tree has its unique encoding from the root to the node.
For example, if we encode $1$ as choosing the left branch and $0$ as choosing the right branch, 
$n_9$'s encoding in tree in Figure~\ref{fig:Model} is $110$ and $n_{15}$'s encoding is $000$.
Denote $b_j(n)$ as the encoding of node $n$ in level $j$.
In hierarchical softmax's formulation, the next-word's probability given the context is derived as
\begin{align} \label{equ:HSProbability}
    P\big(n | context\big) = \prod_{j=1}^w P\big(b = b_j(n) \big| l_j(n), context\big),
\end{align}
where $w$ is the length of leaf node $n$'s encoding, and $l_j(n)$ is $n$'s ancestor node in level $j$.

In such a way, hierarchical softmax solves the probability calculation problem by avoiding the normalization term (each word in the corpus needs to be traversed) 
in conventional softmax.
However, to find the most possible leaf, the model still has to traverse the entire corpus.
Traversing each level's most possible node top-down along the tree path can not guarantee to successfully retrieve the optimal leaf.
Therefore, hierarchical softmax's formulation is not suitable for large-scale retrieval problem.
In addition, according to Equation~\ref{equ:HSProbability}, 
each non-leaf node in tree is trained as a binary classifier to discriminate between its two children nodes.
But if two nodes are neighbors in the tree, they are probably to be similar.
In recommendation scenario, it's likely that user is interested in both two children.
Hierarchical softmax's model focuses on distinguishing optimal and suboptimal choices, 
which may lose the capability of discriminating from a global view.
If greedy beam search is used to retrieve those most possible leaf nodes, 
once bad decisions are made in upper levels of the tree, the model may fail to find relatively better results among those low quality candidates in lower levels.
YouTube's work \cite{Covington2016Deep} also reports that they have tried hierarchical softmax to learn user and item embeddings, 
while it performs worse than sampled-softmax \cite{Jean2014On} manner.

Given that hierarchical softmax's formulation is not suitable for large-scale recommendation, 
we propose a new tree model formulation in the following section.

\subsection{Tree-based Model Formulation} \label{section:Formulation}

To address the problem of efficient top-k retrieval of most preferred items, 
we propose a max-heap like tree probability formulation.
Max-heap like tree is a tree structure where every non-leaf node $n$ in level $j$ satisfies the following equation for each user $u$:
\begin{align} \label{equ:Property}
    P^{(j)}(n | u) = \frac{\max\limits_{n_c \in \{n\text{'s children nodes in level }j+1\}} P^{(j+1)}(n_c | u)}{\alpha^{(j)}},
\end{align}
where $P^{(j)}(n | u)$ is the ground truth probability that user $u$ is interested in $n$.
$\alpha^{(j)}$ is the layer-specific normalization term of level $j$ to ensure that the probability sum in the level equals to 1.
Equation~\ref{equ:Property} says that a parent node's ground truth preference equals to the maximum preference of its children nodes, divided by the normalization term.
Note that we slightly abuse the notation and let $u$ denote a specific user state. 
In other words, a specific user state $u$ may transfer to another state $u'$ once the user has a new behavior.

The goal is to find $k$ leaf nodes with largest preference probabilities.
Suppose that we have each node $n$'s ground truth $P^{(j)}(n | u)$ in the tree, 
we can retrieve $k$ nodes with largest preference probabilities layer-wise, 
and only those children nodes of each level's top $k$ need to be explored.
In this way, top $k$ leaf nodes can be ultimately retrieved.
Actually, we don't need to know each tree node's exact ground truth probability in the above retrieval process.
What we need is the order of the probabilities in each level to help find the top $k$ nodes in the level.
Based on this observation, we use user's implicit feedback data and neural network to train each level's discriminator
that can tell the order of preference probabilities.

Suppose that user $u$ has an interaction with leaf node $n_d$, 
i.e., $n_d$ is a positive sample node for $u$.
It means an order $P^{(m)}(n_d | u) > P^{(m)}(n_t | u)$, 
where $m$ is the level of leaves and $n_t$ is any other leaf node.
In any level $j$, denote $l_j(n_d)$ as $n_d$'s ancestor in level $j$.
According to the formulation of tree in Equation~\ref{equ:Property}, 
we can derive that $P^{(j)}(l_j(n_d) | u) > P^{(j)}(n_q | u)$, 
where $n_q$ is any node in level $j$ except $l_j(n_d)$.
In basis of the above analysis, we can use negative sampling \cite{Mikolov2013Distributed} to train each level's 
order discriminator. In detail, leaf node that have interaction with $u$, and its
ancestor nodes constitute the set of positive samples in each level for $u$. 
And randomly selected nodes except positive ones in each level constitute the set of negative samples.
Those green and red nodes in Figure~\ref{fig:Model} give examples for sampling.
Suppose that given a user and its state, the target node is $n_{13}$. 
Then, $n_{13}$'s ancestors are positive samples, and those randomly sampled red nodes in each level are negative samples.
These samples are then fed into binary probability models to get levels' order discriminators. 
We use one global deep neural network binary model with different input for all levels' order discriminators.
Arbitrary advanced neural network can be adopted to improve model capability.

Denote $\mathcal{Y}_u^+$ and $\mathcal{Y}_u^-$ as the set of positive and negative samples for $u$.
The likelihood function is then derived as: 
\begin{align} \label{equ:Likelihood}
    \prod_u \Big(\prod_{n \in \mathcal{Y}_u^+} P\big(\hat{y}_u(n) = 1 | n, u\big)  \prod_{n \in \mathcal{Y}_u^-} P\big(\hat{y}_u(n) = 0 | n, u\big)\Big),
\end{align}
where $\hat{y}_u(n)$ is the predicted label of node $n$ given $u$.
$P\big(\hat{y}_u(n) | n, u\big)$ is the output of binary probability model, taking user state $u$ and the sampled node $n$ as input.
The corresponding loss function is 
\begin{align} \label{equ:LossFunction}
    & - \sum_u \sum_{n \in \mathcal{Y}_u^+ \cup \mathcal{Y}_u^-} \nonumber \\
    & y_u(n) \log P\big(\hat{y}_u(n) \!=\! 1 | n\!,u\big) \!+\!\big(1 \!-\! y_u(n)\big) \log P\big(\hat{y}_u(n) \!=\! 0 | n\!,u\big),
\end{align}
where $y_u(n)$ is the ground truth label of node $n$ given $u$. 
Details about how to train the model according to the loss function are in Section~\ref{section:NeuralNetworkArchitecture}.

Note that the proposed sampling method is quite different from the underlying one in hierarchical softmax.
Compared to the method used in hierarchical softmax which leads the model to distinguish optimal and suboptimal results, 
we randomly select negative samples in the same level for each positive node.
Such method makes each level's discriminator be an intra-level global one.
Each level's global discriminator can make precise decisions independently, without depending on the goodness of upper levels' decisions.
The global discriminating capability is very important for hierarchical recommendation approaches.
It ensures that even if the model makes bad decision and low quality nodes leak into the candidate set in an upper-level, 
those relatively better nodes rather than very bad ones can be chosen by the model in the following levels.

Given a recommendation tree and an optimized model, the detailed hierarchical prediction algorithm is described in Algorithm~\ref{algo:Retrieve}.
The retrieval process is layer-wise and top-down. Suppose that the desired candidate item number is $k$.
For corpus $C$ with size $|C|$, traversing at most $2 * k * \log |C|$ nodes can get the final recommendation set in a complete binary tree.
The number of nodes need to be traversed is in a logarithmic relation w.r.t. corpus size,
which makes advanced binary probability models possible to be employed. 

\begin{algorithm} 
    \caption{Layer-wise Retrieval Algorithm in Prediction}
    \label{algo:Retrieve}
    \KwIn{User state $u$, the recommendation tree, the desired item number $k$, the learnt model}
    \KwOut{The set of recommended leaf nodes}
    Result set $A = \emptyset$, candidate set $Q = \{\text{the root node }n_1\}$\;
    \Repeat{$|Q| == 0$}{
        If there are leaf nodes in $Q$, remove them from $Q$ and insert them into $A$\;
        Calculate $P\big(\hat{y}_u(n) = 1|n, u\big)$ for each remaining node $n \!\in\! Q$\;
        Sort nodes in $Q$ in descending order of $P\big(\hat{y}_u(n) = 1|n, u\big)$ and derive the set of top $k$ nodes as $I$\;
        $Q = \big\{\text{children nodes of }n  | n \in I\big\}$\;
    }
    Return the top $k$ items in set $A$, according to $P\big(\hat{y}_u(n) = 1|n, u\big)$\;
\end{algorithm}

Our proposed TDM method not only reduces the amount of calculation when making prediction, 
it also has potential to improve recommendation quality compared with brute-force search in all leaf nodes.
Without the tree, training a model to find optimal items directly is a difficult problem because of the corpus size.
Employing the tree hierarchy, a large-scale recommendation problem is divided into many smaller problems.
There only exist a few nodes in high levels of the tree, thus the discrimination problem is easier.
And decisions made by high levels refine the candidate set, which may help lower levels make better judgments.
Experimental results in Section~\ref{section:EmpiricalAnalysis} will show that 
the proposed hierarchical retrieval approach performs better than direct brute-force search.

\subsection{The Deep Model} \label{section:NeuralNetworkArchitecture}
\begin{figure*}[!t]
    \centering
    \includegraphics[height=7in, width=7in,keepaspectratio]{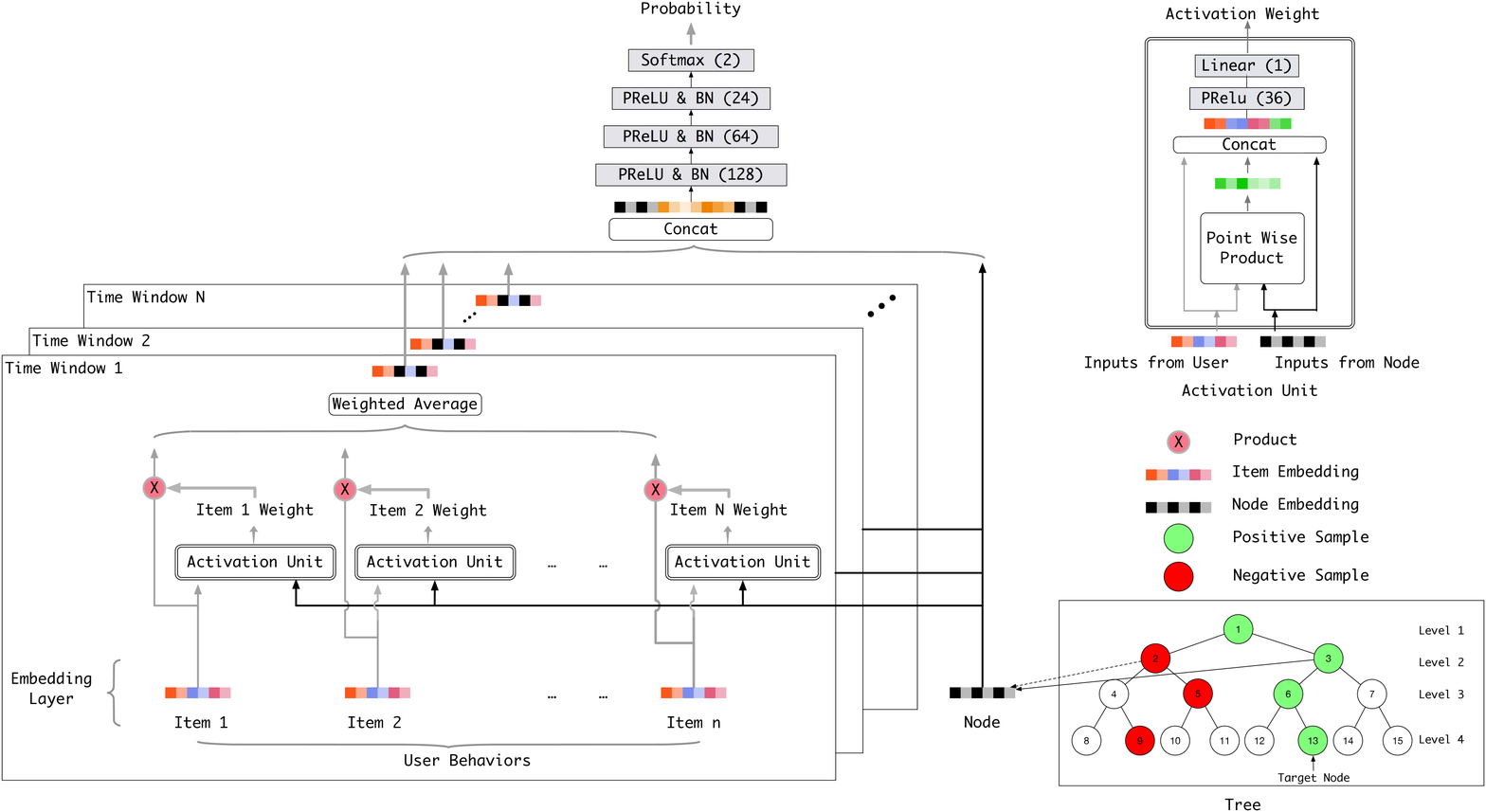}
    \caption{The tree-based deep model architecture. User behaviors are divided into different time windows according to the timestamp.
    In each time window, item embeddings are weighted averaged, and the weights come from activation units. 
    Each time window's output along with the candidate node's embedding are concatenated as the neural network input.
    After three fully connected layers with PReLU \cite{Xu2015Empirical} activation and batch normalization \cite{Ioffe2015Batch}, 
    a binary softmax is used to yield the probability whether the user is interested in the candidate node. 
    Each item and its corresponding leaf node share the same embedding.
    All embeddings are randomly initialized.}
    \label{fig:Model}
\end{figure*}

In the following part, we introduce the deep model we use. The entire model is illustrated in Figure~\ref{fig:Model}.
Inspired by the click-through rate prediction work \cite{Zhou2017Deep},
we learn low dimensional embeddings for each node in the tree, and use attention module to softly searching for related behaviors for better user representation. 
To exploit user behavior that contains timestamp information, 
we design the block-wise input layer to distinguish behaviors that lie in different time windows.
The historical behaviors can be divided into different time windows along the timeline, and item embeddings in each time window is weighted averaged.
Attention module and the following network greatly strengthen the model capability, and also make user's preferences over candidate items can not be regulated to inner product form.

The embeddings of tree nodes and the tree structure itself are also parts of the model. To minimize Loss~\ref{equ:LossFunction},
the sampled nodes and the corresponding features are used to train the network.
Note that we only illustrate the usage of user behavior feature in Figure~\ref{fig:Model} for briefness,
while other features like user profile or contextual feature can be used with no obstacles in practice.

\subsection{Tree Construction and Learning} \label{section:TreeLearning}
The recommendation tree is a fundamental part of the tree-based deep recommendation model.
Unlike multiclass and multi-label classification works \cite{weston2013label, prabhu2014fastxml} where tree is used to partition samples or labels, 
our recommendation tree indexes items for retrieval.
In hierarchical softmax \cite{Morin2005Hierarchical}, the word hierarchy is built according to expert knowledge from WordNet \cite{Lin1999WordNet}.
In the scenario of recommendation, not every corpus can provide specific expert knowledge.
An intuitive alternation is to construct the tree using hierarchical clustering methods in basis of item concurrence or similarity drawn from the dataset.
But the clustered tree may be quite imbalanced, which is detrimental for training and retrieval.
Given pairwise item similarity, algorithm in \cite{Bengio2010Label} gives a way to split items into subsets recursively by spectral clustering \cite{Ng2001On}.
However, spectral clustering is not scalable enough (cubic time complexity w.r.t. corpus size) for large-scale corpus. 
In this section, we focus on reasonable and feasible tree construction and learning approaches.

\paragraph{\textbf{Tree initialization}} 
Since we suppose the tree to represent user interests' hierarchical information, 
it's natural to build the tree in a way that similar items are organized in close positions.
Given that category information is extensive available in many domains, we intuitively come up with a method leveraging item's category information to build the initial tree.
Without loss of generality, we take binary tree as an example in this section.
Firstly, we sort all categories randomly, and place items belonging to the same category together in an intra-category random order.
If an item belongs to more than one category, the item is assigned to a random one for uniqueness.
In such way, we can get a list of ranked items.
Secondly, those ranked items are halved to two equal parts recursively until the current set contains only one item, 
which could construct a near-complete binary tree top-down.
The above kind of category-based initialization can get better hierarchy and results in our experiments than a complete random tree.

\paragraph{\textbf{Tree learning}} 
As a part of the model, each leaf node's embedding can be learnt after model training. Then we use the learnt leaf nodes' embedding 
vectors to cluster a new tree. Considering the corpus size, we use k-means clustering algorithm for its good scalability.
At each step, items are clustered into two subsets according to their
embedding vectors. Note that the two subsets are adjusted to equal for a more balanced tree. The recursion stops when only one item is left, and a binary tree could be constructed in such a top-down way.
In our experiments, it takes about an hour to construct such a cluster tree when the corpus size is about 4 millions, using a single machine.
Experimental results in Section~\ref{section:Experiment} will show the effectiveness of the given tree learning algorithm.

The deep model and tree structure are learnt jointly in an alternative way: 1) Construct an initial tree and train the model till converging; 
2) Learn to get a new tree structure in basis of trained leaf nodes' embeddings; 3) Train the model again with the learnt new tree structure.

\section{Online Serving} \label{section:Serving}

\begin{figure}[H]
    \centering
    \includegraphics[width=1.0\columnwidth]{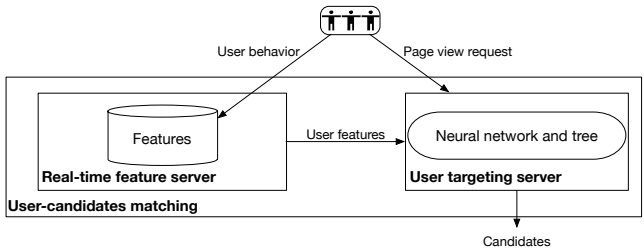}
    \caption{The online serving system of the tree-based model, where user feature is assembled asynchronously.}
    \label{fig:Serving}
\end{figure}

Figure~\ref{fig:Serving} illustrates the online serving system of the proposed method.
Input feature assembling and item retrieval are split into two asynchronous stages.
Each user behavior including click, purchase and adding item into shopping cart will strike the real-time feature server to assemble new input features. 
And once receiving page view request, the user targeting server will use the pre-assembled features to 
retrieve candidates from the tree. As described in Algorithm~\ref{algo:Retrieve}, the retrieval is layer-wise and the trained neural network
is used to calculate the probability that whether a node is preferred given the input features.

\section{EXPERIMENTAL STUDY} \label{section:Experiment}
We study the performance of the proposed tree-based model in this section. 
Experimental results in MovieLens-20M \cite{Harper2015The} and Taobao advertising dataset called UserBehavior are presented.
In the experiments, we compare the proposed method to other existing methods to show the effectiveness of the model,
and empirical study results show how the tree-based model and tree learning algorithm work.

\subsection{Datasets}
The experiments are conducted in two large-scale real-world datasets with timestamps: 1) users' movie viewing data from MovieLens \cite{Harper2015The};
2) a user-item behavior dataset from Taobao called UserBehavior. In more details:

\textbf{MovieLens-20M:} It contains user-movie ratings with timestamps in this dataset. As we deal with
implicit feedback problem, the ratings are binarized by keeping the ratings of four or higher, which is 
a common way in other works \cite{Liang2016Factorization, Devooght2016Collaborative}. Besides, only the users 
who have watched at least 10 movies are kept. To create training, validation and testing sets, we randomly sample 
$1,000$ users as testing set and another $1,000$ users as validation set, while the rest users constitute the training set \cite{Devooght2016Collaborative}.
For validation and testing sets, the first half of user-movie views along the timeline is regarded as known behaviors to
predict the latter half.

\textbf{UserBehavior\footnote{\url{https://tianchi.aliyun.com/datalab/dataSet.html?spm=5176.100073.0.0.614435eeJVooEG&dataId=649}}:} This dataset is a subset of Taobao user behavior data. 
We randomly select about $1$ million users who have 
behaviors including click, purchase, adding item to shopping cart and item favoring during November 25 to December 03, 2017. 
The data is organized in a very similar form to MovieLens-20M, i.e., a user-item behavior consists of user ID, item ID, item's category ID, behavior type and timestamp.
As we do in MovieLens-20M, only the users who have at least 10 behaviors are kept. $10,000$ users are randomly selected as testing set and another randomly selected $10,000$ users are validation set.
Items' categories are from the bottom level of Taobao's current commodity taxonomy.
Table~\ref{table:Dataset} summarizes the major dimensions of the above two datasets after preprocessing.

\begin{table}[!htb]
    \small
    \begin{tabular}{ccc}
        \toprule[1pt]
                         & \textbf{MovieLens-20M} & \textbf{UserBehavior} \\
        \hline
        \# of users      & 129,797                & 969,529               \\
        \# of items      & 20,709                 & 4,158,142             \\
        \# of categories & 20                     & 9,436                 \\
        \# of records    & 9,939,873              & 100,020,395           \\
        \bottomrule[1pt]
    \end{tabular}
    \caption{Dimensions of the two datasets after preprocessing. One record is a user-item pair that 
    represents user feedback.}
    \label{table:Dataset}
\end{table}

\subsection{Metrics and Comparison Methods}
To evaluate the effectiveness of different methods, we use Precision@M, Recall@M and F-Measure@M metrics \cite{Liang2016Factorization}.
Derive the recalled set of items for a user $u$ as $\mathcal{P}_u$ (|$\mathcal{P}_u| = M$) and the user's ground truth set as $\mathcal{G}_u$. Precision@M
and Recall@M are
\begin{align} \label{equ:Precision}
    Precision@M(u) = \frac{|\mathcal{P}_u \cap \mathcal{G}_u|}{M}, Recall@M(u) = \frac{|\mathcal{P}_u \cap \mathcal{G}_u|}{|\mathcal{G}_u|},
\end{align}
and F-Measure@M is 
\begin{align} \label{equ:FMeasure}
    F\text{-}Measure@M(u) = \frac{2 * Precision@M(u) * Recall@M(u)}{ Precision@M(u) + Recall@M(u)}.
\end{align}

As we emphasize, recommendation results' novelty is responsible for user experience. 
Existing work \cite{Castells2011Novelty} gives several approaches to measure the novelty of recommended list of items.
Following one of its definition, the Novelty@M is defined as 
\begin{align} \label{equ:Probability}
    Novelty@M(u) = \frac{|\mathcal{P}_u \setminus \mathcal{S}_u|}{M},
\end{align}
where $\mathcal{S}_u$ is the set of items that have interactions with user $u$ before recommending.
User average of the above four metrics in testing set are used to compare the following methods:

\begin{itemize}
    \item \textbf{FM}\cite{Rendle2010Factorization}. FM is a framework for factorization tasks.
    We use the implementation of FM provided by xLearn\footnote{\url{https://github.com/aksnzhy/xlearn}} project.
    \item \textbf{BPR-MF}\cite{rendle2009bpr}. We use its matrix factorization form for implicit feedback recommendation.
    Implementation of BPR-MF provided by \cite{gantner2011mymedialite} is used. 
    \item \textbf{Item-CF}\cite{Sarwar2001Item}. Item-based collaborative filtering is one of the most widely 
    used personalized recommendation method in production with large-scale corpus \cite{linden2003amazon}. 
    It's also one of the major candidate generation approaches in Taobao.
    We use the implementation of item-CF provided by Alibaba machine learning platform.
    \item \textbf{YouTube product-DNN}\cite{Covington2016Deep} is the deep recommendation
    approach proposed by YouTube. Sampled-softmax \cite{Jean2014On} is employed in training, and the inner product of user and item's embeddings reflects the preference. 
    We implement YouTube product-DNN in Alibaba deep learning platform with the same input features with our proposed model.
    Exact kNN search in inner product space is adopted in prediction.
    \item \textbf{TDM attention-DNN} (tree-based deep model using attention network) is our proposed method in Figure~\ref{fig:Model}.
    The tree is initialized in the way described in Section~\ref{section:TreeLearning} and keeps unchanged during the experiments. The implementation is available in GitHub\footnote{\url{https://github.com/alibaba/x-deeplearning/tree/master/xdl-algorithm-solution/TDM}}.
\end{itemize}

For FM, BPR-MF and item-CF, we tune several most important hyper-parameters based on the validation set, 
i.e., the number of factors and iterations in FM and BPR-MF, the number of neighbors in item-CF. 
FM and BPR-MF require that the users in testing or validation set also have feedback in training set. 
Therefore, we add the first half of user-item interactions along the 
timeline in testing and validation set into the training set in both datasets.
For YouTube product-DNN and TDM attention-DNN,
the node embeddings' dimension is set to 24, because a higher dimension doesn't perform significantly better in our experiments.
The hidden unit numbers of three fully connected layers are 128, 64 and 24 respectively. 
According to the timestamp, user behaviors are divided into 10 time windows.
In YouTube product-DNN and TDM attention-DNN, for each implicit feedback we randomly select 100 negative samples in MovieLens-20M and 600 negative samples in UserBehavior.
Note that the negative sample number of TDM is the sum of all levels.
And we sample more negatives for levels near to leaf.

\subsection{Comparison Results} \label{section:ComparisonResult}
The comparison results of different methods are shown in Table~\ref{table:ComparisonResults} above the dash line. 
Each metric is the average across all the users in testing set, and the presented values are the average across five different runs for methods with variance.

\begin{table*}[!htb]
    \small
    \addtolength{\tabcolsep}{0pt} 
    \centering 
    \begin{tabular}{c|c|cccc|cccc}
        \Xhline{1.0pt}
        \multirow{2}{30pt}{\centering Filtering} & \multirow{2}{30pt}{\centering Method} & \multicolumn{4}{c|}{MovieLens-20M (@10)} & \multicolumn{4}{c}{UserBehavior (@200)} \\
        \cline{3-10}
                                               &                     & Precision        & Recall           & F-Measure        & Novelty           & Precision       & Recall           & F-Measure       & Novelty           \\
        \hline
        \multirow{8}{*}{None}          & FM                  & 8.35\%           & 5.12\%           & 5.03\%           & 70.76\%           & 0.31\%          & 1.67\%           & 0.45\%          & \textbf{99.58\%}  \\
                                               & BPR-MF              & 8.10\%           & 5.09\%           & 5.02\%           & 62.56\%           & 0.44\%          & 1.84\%           & 0.64\%          & 99.56\%           \\
                                               & Item-CF             & 8.25\%           & 5.66\%           & 5.29\%           & 59.46\%           & 1.47\%          & 6.95\%           & 2.18\%          & 97.07\%           \\
                                               & YouTube product-DNN & 11.87\%          & 8.71\%           & 7.96\%           & 71.38\%           & 1.48\%          & 7.58\%           & 2.23\%          & 98.48\%           \\
                                               & TDM attention-DNN   & \textbf{14.06\%} & \textbf{10.55\%} & \textbf{9.49\%}  & \textbf{74.15\%}  & \textbf{2.00\%} & \textbf{10.81\%} & \textbf{3.03\%} & 97.30\%           \\
        \cdashline{2-10}[1pt/2pt] 
                                               & TDM product-DNN     & 12.20\%          & 9.18\%           & 8.23\%           & 72.78\%           & 1.50\%          & 7.80\%           & 2.26\%          & 98.36\%           \\
                                               & TDM DNN             & 13.35\%          & 10.10\%          & 8.98\%           & 72.18\%           & 1.78\%          & 9.67\%           & 2.70\%          & 97.94\%           \\
                                               & TDM attention-DNN-HS& 10.92\%          & 9.16\%           & 7.94\%           & 81.00\%           & 1.47\%          & 8.20\%           & 2.25\%          & 98.28\%           \\
        \hline
        \multirow{8}{*}{Interacted items} & FM                & 13.39\%          & 6.87\%           & 7.10\%           & \textbf{100.00\%} & 0.11\%          & 0.56\%           & 0.17\%          & \textbf{100.00\%} \\
                                               & BPR-MF              & 13.39\%          & 6.95\%           & 7.17\%           & \textbf{100.00\%} & 0.36\%          & 1.51\%           & 0.53\%          & \textbf{100.00\%} \\
                                               & Item-CF             & 15.61\%          & 8.86\%           & 8.81\%           & \textbf{100.00\%} & 0.68\%          & 4.38\%           & 1.06\%          & \textbf{100.00\%} \\
                                               & YouTube product-DNN & 16.51\%          & 10.70\%          & 10.04\%          & \textbf{100.00\%} & 0.93\%          & 5.67\%           & 1.44\%          & \textbf{100.00\%} \\
                                               & TDM attention-DNN   & \textbf{17.77\%} & \textbf{12.31\%} & \textbf{11.33\%} & \textbf{100.00\%} & \textbf{1.16\%} & \textbf{7.50\%}  & \textbf{1.81\%} & \textbf{100.00\%} \\
        \cdashline{2-10}[1pt/2pt] 
                                               & TDM product-DNN     & 17.29\%          & 11.87\%          & 10.91\%          & 100.00\%          & 0.92\%          & 5.68\%           & 1.44\%          & 100.00\%          \\
                                               & TDM DNN             & 17.82\%          & 12.12\%          & 11.31\%          & 100.00\%          & 1.02\%          & 6.97\%           & 1.68\%          & 100.00\%          \\
                                               & TDM attention-DNN-HS& 14.06\%          & 10.72\%          & 9.58\%           & 100.00\%          & 0.86\%          & 5.79\%           & 1.36\%          & 100.00\%          \\
        \Xhline{1.0pt}
    \end{tabular}
    \caption{The comparison results of different methods in MovieLens-20M and UserBehavior datasets. According to the different corpus size, 
    metrics are evaluated @10 in MovieLens-20 and @200 in UserBehavior. In experiments of filtering interacted items, 
    the recommendation results and ground truth only contain items that the user has not yet interacted with before.}
    \label{table:ComparisonResults}
    \normalsize
\end{table*}

First, the results indicate that the proposed TDM attention-DNN outperforms all the baselines significantly in both datasets on most of the metrics.
Comparing to the second best YouTube product-DNN approach, TDM attention-DNN achieves $21.1\%$ and $42.6\%$ improvements on recall metric in two datasets respectively without filtering.
This result proves the effectiveness of advanced neural network and hierarchical tree search adopted by TDM attention-DNN.
Among the methods that model user preference over items in inner product form, 
YouTube product-DNN outperforms BPR-MF and FM because of the usage of neural network.
The widely used item-CF method gets worst novelty results, since it has strong memories about what the user has already interacted.

To improve the novelty, a common way in practice is to filter those interacted items in recommendation set \cite{Liang2016Factorization, Devooght2016Collaborative}, 
i.e., only those novel items could be ultimately recommended.
Thus, it's more important to compare accuracy in a complete novel result set. 
In this experiment, the result set size will be complemented to required number $M$ if its size is smaller than $M$ after filtering.
The bottom half of Table~\ref{table:ComparisonResults} shows that TDM attention-DNN outperforms all baselines in large margin as well after filtering interacted items.

To further evaluate the exploration ability of different methods, we do experiments by excluding those interacted categories from recommendation results.
Results of each method are also complemented to satisfy the size requirement. 
Indeed, category-level novelty is currently the most important novelty metric in Taobao recommender system, 
as we want to reduce the amount of recommendations similar to user's interacted items.
Since MovieLens-20M has only 20 categories in total, these experiments are only conducted in UserBehavior dataset and results are shown in Table~\ref{table:Novelty}.
Take the recall metric for example. We can observe that item-CF's recall is only $1.06\%$, 
because its recommendation results can hardly jump out of user's historical behaviors.
YouTube product-DNN gets much better results compared to item-CF, since it can explore user's potential interests from the entire corpus.
The proposed TDM attention-DNN performs $34.3\%$ better in recall than YouTube's inner product manner. 
Such huge improvement is very meaningful for recommender systems, and it proves that more advanced model is an enormous difference for recommendation problem.

\begin{table}[!htb]
    \small
    \begin{tabular}{cccc}
        \toprule[1pt]
        Method (@200)              & Precision   & Recall      & F-Measure   \\
        \hline
        Item-CF            & 0.07\%          & 1.06\%          & 0.13\%          \\
        YouTube product-DNN & 0.26\%          & 3.09\%          & 0.45\%          \\
        TDM attention-DNN   & \textbf{0.35\%} & \textbf{4.15\%} & \textbf{0.60\%} \\
        \bottomrule[1pt]
    \end{tabular}
    \caption{Results in UserBehavior dataset. Items belong to interacted categories are excluded from recommendation results and ground truth.}
    \label{table:Novelty}
\end{table}

\subsection{Empirical Analysis} \label{section:EmpiricalAnalysis}
\paragraph{\textbf{Variants of TDM}} To comprehend the proposed TDM method itself, 
we derive and evaluate several variants of TDM:
\begin{itemize}
    \item \textbf{TDM product-DNN.} To find out whether advanced neural network can benefit the results in TDM, 
    we test the variant TDM product-DNN. TDM product-DNN uses the same inner product manner as YouTube product-DNN.
    Specifically, the attention module in Figure~\ref{fig:Model} is removed, and the node embedding term is also removed from the network input.
    The inner product of node embedding and the third fully connected layer's output (without PReLU and BN) along with a sigmoid activation constitute the new binary classifier.
    \item \textbf{TDM DNN.} To further verify the improvements brought by attention module in TDM attention-DNN, 
    we test the variant TDM DNN that only removes the activation unit, i.e., all items' weights are $1.0$ in Figure~\ref{fig:Model}.
    \item \textbf{TDM attention-DNN-HS.} As mentioned in Section~\ref{section:Mechanism}, hierarchical softmax (HS) method \cite{Morin2005Hierarchical} is not suitable for recommendation.
    We test the TDM attention-DNN-HS variant, i.e., use positive nodes' neighbors as negative samples instead of randomly selected ones.
    Correspondingly, in retrieval of Algorithm~\ref{algo:Retrieve}, the ranking indicator changes from a single node's $P\big(\hat{y}_u(n) = 1|n, u\big)$ to $\prod_{n' \in \text{ n's ancestors}}P\big(\hat{y}_u(n') = 1|n', u\big)$.
    Attention-DNN is used as the network structure.
\end{itemize}

The experimental results of the above variants in both datasets are shown in Table~\ref{table:ComparisonResults} under the dash line.
Comparing TDM attention-DNN to TDM DNN, the near $10\%$ recall improvement in UserBehavior dataset indicates that the attention module takes impressive efforts.
TDM product-DNN performs worse than TDM DNN and TDM attention-DNN, since the inner product manner is much less powerful than the neural network interaction form.
These results prove that introducing advanced models in TDM can significantly improve the recommendation performance.
Note that TDM attention-DNN-HS gets much worse results compared to TDM attention-DNN, since hierarchical softmax's formulation doesn't fit for recommendation problem.

\begin{figure}[tb]
    \centering
    \includegraphics[width=0.6\columnwidth]{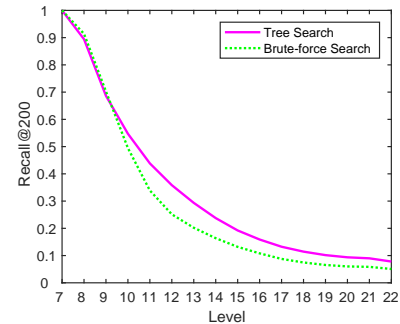}
    \caption{The results of layer-wise Recall@200 in UserBehavior dataset. 
    The ground truth in testing set is traced back to each node's ancestors, till the root node.}
    \label{fig:LayerWise}
\end{figure}

\paragraph{\textbf{Role of the tree}}
Tree is the key component of the proposed TDM method. It not only acts as an index used in retrieval, but also models the corpus in coarse-to-fine hierarchy.
Section~\ref{section:Formulation} mentioned that directly making fine-grained recommendation is more difficult than a hierarchical way. 
We conduct experiments to prove the point of view.
Figure~\ref{fig:LayerWise} illustrates the layer-wise Recall@200 of hierarchical tree search (Algorithm~\ref{algo:Retrieve}) and 
brute-force search (traverse all nodes in the corresponding level).
The experiments are conducted in UserBehavior dataset with TDM product-DNN model, because it's the only variant that is possible to employ brute-force search.
Brute-force search slightly outperforms tree search in high levels (level 8, 9), since the node numbers there are small.
Once the node number in a level grows, tree search gets better recall results compared to brute-force search,
because the tree search can exclude those low quality results in high levels,
which reduces the difficulty of the problems in low levels.
This result indicates that the hierarchy information contained in the tree structure can help improve recommendation preciseness.

\begin{figure}[tb]
    \centering
    \subfigure[Test Loss]{
        \includegraphics[width=0.22\textwidth]{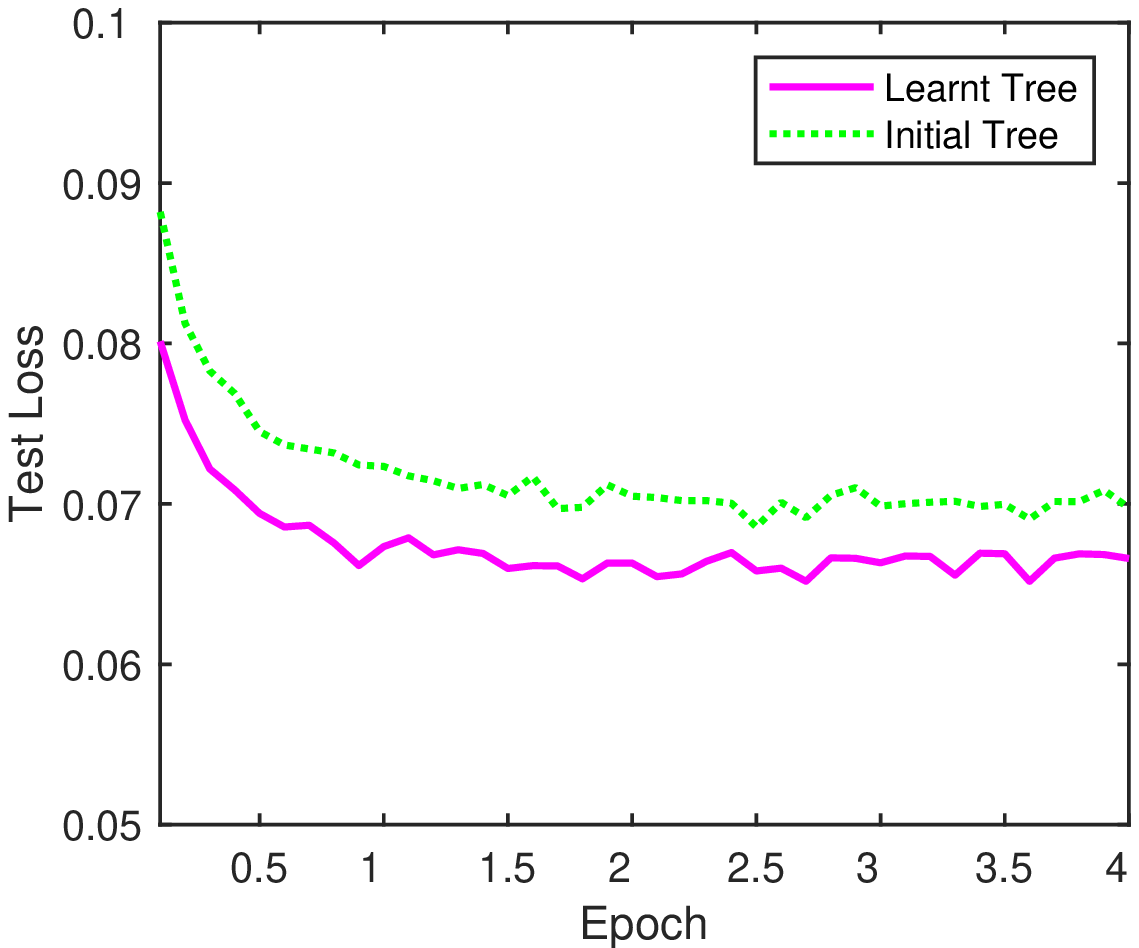}
        \label{fig:LossCurve}
    }
    \subfigure[Test Recall]{
        \includegraphics[width=0.22\textwidth]{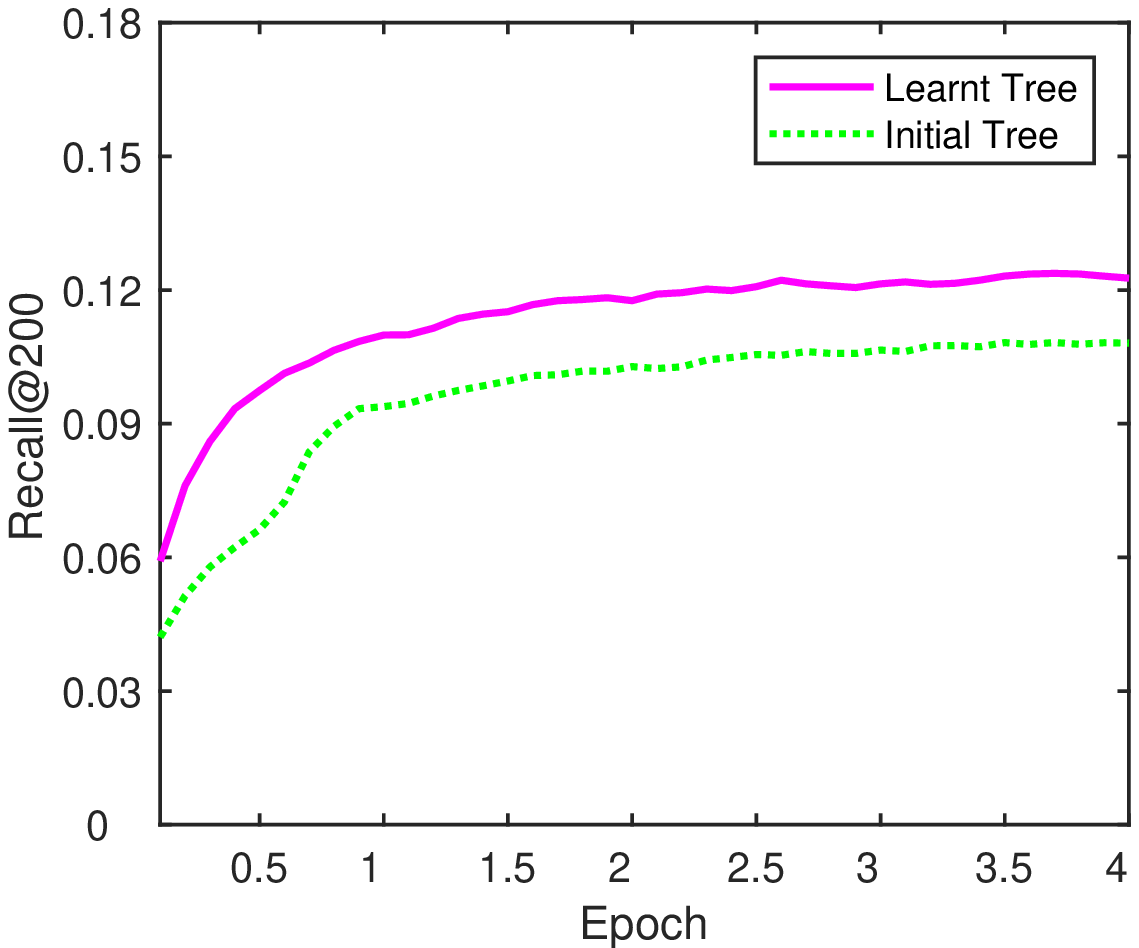}
        \label{fig:RecallCurve}
    }
    \caption{The test loss and test Recall@200 on UserBehavior dataset for initial and learnt tree.}
    \label{fig:TreeLearning}
\end{figure}

\paragraph{\textbf{Tree learning}} In Section~\ref{section:TreeLearning}, 
we propose the tree initialization and learning algorithms.
Table~\ref{table:TreeLearning} gives the comparison results between initial tree and learnt tree.
From the results, we can observe that the trained model with learnt tree structure significantly outperforms the initial one.
For example, the recall metric of learnt tree increases from $4.15\%$ to $4.82\%$ compared to initial tree in experiments of filtering interacted categories, 
which surpasses YouTube product-DNN's $3.09\%$ and item-CF's $1.06\%$ in very large margin.
To further compare these two trees, we illustrate the test loss and recall curve of TDM attention-DNN method w.r.t. training iterations in Figure~\ref{fig:TreeLearning}.
From Figure~\ref{fig:LossCurve}, we can see that the learnt tree structure gets smaller test loss. 
And both Figure~\ref{fig:LossCurve} and \ref{fig:RecallCurve} indicate that the model converges to better results with learnt tree.
The above results prove that the tree learning algorithm can improve the hierarchy of items, further to facilitate training and prediction.

\begin{table}[!htb]
    \small
    \addtolength{\tabcolsep}{0pt} 
    \centering 
    \begin{tabular}{c|c|cccc}
        \Xhline{1.0pt}
        Filtering                                     & Tree & Precision       & Recall           & F-Measure       & Novelty           \\
        \hline
        \multirow{2}{*}{\makecell{None}}                 & Initial    & 2.00\%          & 10.81\%          & 3.03\%          & \textbf{97.30}\%  \\
                                                      & Learnt     & \textbf{2.34\%} & \textbf{12.37\%} & \textbf{3.54\%} & 96.68\%           \\
        \hline
        \multirow{2}{*}{\makecell{Interacted \\ items}}      & Initial    & 1.16\%          & 7.50\%           & 1.81\%          & \textbf{100.00\%} \\
                                                      & Learnt     & \textbf{1.33\%} & \textbf{8.38\%}  & \textbf{2.09\%} & \textbf{100.00\%} \\
        \hline
        \multirow{2}{*}{\makecell{Interacted \\ categories}} & Initial    & 0.35\%          & 4.15\%           & 0.60\%          & \textbf{100.00\%} \\
                                                      & Learnt     & \textbf{0.40\%} & \textbf{4.82\%}  & \textbf{0.69\%} & \textbf{100.00\%} \\
        \Xhline{1.0pt}
    \end{tabular}
    \caption{Comparison results of different tree structures in UserBehavior dataset using TDM attention-DNN model (@200). Tree is initialized and learnt according to the algorithm described in Section~\ref{section:TreeLearning}.}
    \label{table:TreeLearning}
    \normalsize
\end{table}

\subsection{Online Results} \label{section:OnlineResults}
We evaluate the proposed TDM method in Taobao display advertising platform with real traffic.
The experiments are conducted in \emph{Guess What You Like} column of Taobao App Homepage.
Two online metrics are used to measure the performance: click-through rate (CTR) and revenue per mille (RPM).
Details are as follows:
\begin{align} \label{equ:OnlineMetrics}
    CTR = \frac{\text{\# of clicks}}{\text{\# of impressions}},RPM = \frac{\text{Ad revenue}}{\text{\# of impressions}} * 1000.
\end{align}

In our advertising system, advertisers bid on some given ad clusters.
There are about $1.4$ million clusters and each ad cluster contains hundreds or thousands of similar ads.
The experiments are conducted in the granularity of ad cluster to keep consistent with the existing system.
The comparison method is mixture of logistic regression \cite{gai2017learning} that
used to pick out superior results only from those interacted clusters, which is a strong baseline.
Since there are many stages in the system like CTR prediction \cite{Zhou2017Deep, ge2017image} and ranking \cite{OCPC} as illustrated in Figure~\ref{fig:Architechture},
deploying and evaluating the proposed TDM method online is a huge project, which involves the linkage and optimization of the whole system.
We have finished the deployment of the first TDM DNN version so far and evaluated its improvements online.
Each of the comparison buckets has 5\% of all online traffic.
It's worth mentioning that there are several online simultaneously running recommendation methods.
They take efforts in different point of views, and their recommendation results are merged together for the following stages.
TDM only replaces the most effective one of them while keeping other modules unchanged.
The average metric lift rates of the testing bucket with TDM are listed in Table~\ref{table:OnlineResults}.

\begin{table}[!htb]
    \begin{tabular}{p{50pt}p{50pt}p{18pt}}
        \toprule[1pt]
        Metric & CTR &RPM\\
        \hline
        Lift Rate & 2.1\% &6.4\%  \\
        \bottomrule[1pt]
    \end{tabular}
    \caption{Online results from Jan 22 to Jan 28, 2018 in \emph{Guess What You Like} column of Taobao App Homepage.}
    \label{table:OnlineResults}
\end{table}
As shown in Table~\ref{table:OnlineResults}, the CTR of TDM method increases $2.1\%$. 
This improvement indicates that the proposed method can recall more accurate results for users.
And on the other hand the RPM metric increases $6.4\%$, 
which means the TDM method can also bring more revenue for Taobao advertising platform. 
TDM has been deployed to serve major online traffic,
we believe that the above improvement is only a preliminary result in a huge project, 
and there has room for further improvements.
\paragraph{\textbf{Prediction efficiency}}
TDM makes advanced neural network feasible to interact user and items in large-scale recommendation, 
which opens a new perspective of view in recommender systems.
It's worth mentioning that though advanced neural networks need more calculation when inferring,
but the complexity of a whole prediction process is no larger than $O(k*\log |C| * t)$, 
where $k$ is the required results size, $|C|$ is the corpus size and $t$ is the complexity of network's single feed-forward pass.
This complexity upper bound is acceptable under current CPU/GPU hardware conditions,
and user side's features are shared across different nodes in one retrieval and some calculation could be shared according to model designs. 
In Taobao display advertising system, it actually takes the deployed TDM DNN model about 6 milliseconds to recommend once in average.
Such running time is shorter than the following click-through rate prediction module, 
and is not the system's bottleneck.

\section{Conclusion}\label{section:Conclusion}
We figure out the main challenge for model-based methods to generate recommendations from large-scale corpus, 
i.e., the amount of calculation problem when making prediction.
A tree-based approach is proposed, where arbitrary advanced models can be employed in large-scale recommendation to infer user interests coarse-to-fine along the tree.
Besides training the model, a tree structure learning approach is used, which proves that a better tree structure can lead to significantly better results.
A possible future direction is to design more elaborate tree learning approaches.
We conduct extensive experiments which validate the effectiveness of the 
proposed method, both in recommendation accuracy and novelty. In addition, empirical analysis showcases how and why the proposed method works.
In Taobao display advertising platform, 
the proposed TDM method has been deployed in production, which improves both business benefits and user experience.

\section*{Acknowledgements}
We deeply appreciate Jian Xu, Chengru Song, Chuan Yu, Guorui Zhou and Yongliang Wang for their helpful suggestions and discussions.
Thank Huimin Yi, Yang Zheng, Zelin Hu, Sui Huang, Yin Yang and Bochao Liu for implementing the key components of the training and serving infrastructure.
Thank Haiyang He, Yangyang Fu and Yang Wang for necessary engineering supports.

\bibliographystyle{ACM-Reference-Format}
\bibliography{deepmatch} 


\begin{thebibliography}{00}


\ifx \showCODEN    \undefined \def \showCODEN     #1{\unskip}     \fi
\ifx \showDOI      \undefined \def \showDOI       #1{#1}\fi
\ifx \showISBNx    \undefined \def \showISBNx     #1{\unskip}     \fi
\ifx \showISBNxiii \undefined \def \showISBNxiii  #1{\unskip}     \fi
\ifx \showISSN     \undefined \def \showISSN      #1{\unskip}     \fi
\ifx \showLCCN     \undefined \def \showLCCN      #1{\unskip}     \fi
\ifx \shownote     \undefined \def \shownote      #1{#1}          \fi
\ifx \showarticletitle \undefined \def \showarticletitle #1{#1}   \fi
\ifx \showURL      \undefined \def \showURL       {\relax}        \fi
\providecommand\bibfield[2]{#2}
\providecommand\bibinfo[2]{#2}
\providecommand\natexlab[1]{#1}
\providecommand\showeprint[2][]{arXiv:#2}

\bibitem[\protect\citeauthoryear{Agrawal, Gupta, Prabhu, and Varma}{Agrawal
  et~al\mbox{.}}{2013}]%
        {agrawal2013multi}
\bibfield{author}{\bibinfo{person}{Rahul Agrawal}, \bibinfo{person}{Archit
  Gupta}, \bibinfo{person}{Yashoteja Prabhu}, {and} \bibinfo{person}{Manik
  Varma}.} \bibinfo{year}{2013}\natexlab{}.
\newblock \showarticletitle{Multi-label learning with millions of labels:
  Recommending advertiser bid phrases for web pages}. In
  \bibinfo{booktitle}{{\em Proceedings of the 22nd international conference on
  World Wide Web}}. ACM, \bibinfo{pages}{13--24}.
\newblock


\bibitem[\protect\citeauthoryear{Bengio, Weston, and Grangier}{Bengio
  et~al\mbox{.}}{2010}]%
        {Bengio2010Label}
\bibfield{author}{\bibinfo{person}{Samy Bengio}, \bibinfo{person}{Jason
  Weston}, {and} \bibinfo{person}{David Grangier}.}
  \bibinfo{year}{2010}\natexlab{}.
\newblock \showarticletitle{Label embedding trees for large multi-class tasks}.
  In \bibinfo{booktitle}{{\em International Conference on Neural Information
  Processing Systems}}. \bibinfo{pages}{163--171}.
\newblock


\bibitem[\protect\citeauthoryear{Beygelzimer, Langford, and
  Ravikumar}{Beygelzimer et~al\mbox{.}}{2007}]%
        {Beygelzimer2007Multiclass}
\bibfield{author}{\bibinfo{person}{Alina Beygelzimer}, \bibinfo{person}{John
  Langford}, {and} \bibinfo{person}{Pradeep Ravikumar}.}
  \bibinfo{year}{2007}\natexlab{}.
\newblock \showarticletitle{Multiclass classification with filter trees}.
\newblock \bibinfo{journal}{{\em Gynecologic Oncology\/}}
  \bibinfo{volume}{105}, \bibinfo{number}{2} (\bibinfo{year}{2007}),
  \bibinfo{pages}{312--320}.
\newblock


\bibitem[\protect\citeauthoryear{Castells, Vargas, and Wang}{Castells
  et~al\mbox{.}}{2011}]%
        {Castells2011Novelty}
\bibfield{author}{\bibinfo{person}{Pablo Castells}, \bibinfo{person}{Saúl
  Vargas}, {and} \bibinfo{person}{Jun Wang}.} \bibinfo{year}{2011}\natexlab{}.
\newblock \showarticletitle{Novelty and Diversity Metrics for Recommender
  Systems: Choice, Discovery and Relevance}.
\newblock \bibinfo{journal}{{\em In Proceedings of International Workshop on
  Diversity in Document Retrieval\/}} (\bibinfo{year}{2011}),
  \bibinfo{pages}{29--37}.
\newblock


\bibitem[\protect\citeauthoryear{Cheng, Koc, Harmsen, Shaked, Chandra, Aradhye,
  Anderson, Corrado, Chai, Ispir, et~al\mbox{.}}{Cheng et~al\mbox{.}}{2016}]%
        {cheng2016wide}
\bibfield{author}{\bibinfo{person}{Heng-Tze Cheng}, \bibinfo{person}{Levent
  Koc}, \bibinfo{person}{Jeremiah Harmsen}, \bibinfo{person}{Tal Shaked},
  \bibinfo{person}{Tushar Chandra}, \bibinfo{person}{Hrishi Aradhye},
  \bibinfo{person}{Glen Anderson}, \bibinfo{person}{Greg Corrado},
  \bibinfo{person}{Wei Chai}, \bibinfo{person}{Mustafa Ispir}, {et~al\mbox{.}}}
  \bibinfo{year}{2016}\natexlab{}.
\newblock \showarticletitle{Wide \& deep learning for recommender systems}. In
  \bibinfo{booktitle}{{\em Proceedings of the 1st Workshop on Deep Learning for
  Recommender Systems}}. ACM, \bibinfo{pages}{7--10}.
\newblock


\bibitem[\protect\citeauthoryear{Choromanska and Langford}{Choromanska and
  Langford}{2015}]%
        {Choromanska2015Logarithmic}
\bibfield{author}{\bibinfo{person}{Anna~E Choromanska} {and}
  \bibinfo{person}{John Langford}.} \bibinfo{year}{2015}\natexlab{}.
\newblock \showarticletitle{Logarithmic time online multiclass prediction}. In
  \bibinfo{booktitle}{{\em Advances in Neural Information Processing Systems}}.
  \bibinfo{pages}{55--63}.
\newblock


\bibitem[\protect\citeauthoryear{Covington, Adams, and Sargin}{Covington
  et~al\mbox{.}}{2016}]%
        {Covington2016Deep}
\bibfield{author}{\bibinfo{person}{Paul Covington}, \bibinfo{person}{Jay
  Adams}, {and} \bibinfo{person}{Emre Sargin}.}
  \bibinfo{year}{2016}\natexlab{}.
\newblock \showarticletitle{Deep Neural Networks for YouTube Recommendations}.
  In \bibinfo{booktitle}{{\em ACM Conference on Recommender Systems}}.
  \bibinfo{pages}{191--198}.
\newblock


\bibitem[\protect\citeauthoryear{Devooght and Bersini}{Devooght and
  Bersini}{2016}]%
        {Devooght2016Collaborative}
\bibfield{author}{\bibinfo{person}{Robin Devooght} {and}
  \bibinfo{person}{Hugues Bersini}.} \bibinfo{year}{2016}\natexlab{}.
\newblock \showarticletitle{Collaborative filtering with recurrent neural
  networks}.
\newblock \bibinfo{journal}{{\em arXiv preprint arXiv:1608.07400\/}}
  (\bibinfo{year}{2016}).
\newblock


\bibitem[\protect\citeauthoryear{Gai, Zhu, Li, Liu, and Wang}{Gai
  et~al\mbox{.}}{2017}]%
        {gai2017learning}
\bibfield{author}{\bibinfo{person}{Kun Gai}, \bibinfo{person}{Xiaoqiang Zhu},
  \bibinfo{person}{Han Li}, \bibinfo{person}{Kai Liu}, {and}
  \bibinfo{person}{Zhe Wang}.} \bibinfo{year}{2017}\natexlab{}.
\newblock \showarticletitle{Learning Piece-wise Linear Models from Large Scale
  Data for Ad Click Prediction}.
\newblock \bibinfo{journal}{{\em arXiv preprint arXiv:1704.05194\/}}
  (\bibinfo{year}{2017}).
\newblock


\bibitem[\protect\citeauthoryear{Gantner, Rendle, Freudenthaler, and
  Schmidt-Thieme}{Gantner et~al\mbox{.}}{2011}]%
        {gantner2011mymedialite}
\bibfield{author}{\bibinfo{person}{Zeno Gantner}, \bibinfo{person}{Steffen
  Rendle}, \bibinfo{person}{Christoph Freudenthaler}, {and}
  \bibinfo{person}{Lars Schmidt-Thieme}.} \bibinfo{year}{2011}\natexlab{}.
\newblock \showarticletitle{MyMediaLite: A free recommender system library}. In
  \bibinfo{booktitle}{{\em Proceedings of the fifth ACM conference on
  Recommender systems}}. ACM, \bibinfo{pages}{305--308}.
\newblock


\bibitem[\protect\citeauthoryear{Ge, Zhao, Zhou, Chen, Liu, Yi, Hu, Liu, Sun,
  Liu, et~al\mbox{.}}{Ge et~al\mbox{.}}{2017}]%
        {ge2017image}
\bibfield{author}{\bibinfo{person}{Tiezheng Ge}, \bibinfo{person}{Liqin Zhao},
  \bibinfo{person}{Guorui Zhou}, \bibinfo{person}{Keyu Chen},
  \bibinfo{person}{Shuying Liu}, \bibinfo{person}{Huiming Yi},
  \bibinfo{person}{Zelin Hu}, \bibinfo{person}{Bochao Liu},
  \bibinfo{person}{Peng Sun}, \bibinfo{person}{Haoyu Liu}, {et~al\mbox{.}}}
  \bibinfo{year}{2017}\natexlab{}.
\newblock \showarticletitle{Image Matters: Jointly Train Advertising CTR Model
  with Image Representation of Ad and User Behavior}.
\newblock \bibinfo{journal}{{\em arXiv preprint arXiv:1711.06505\/}}
  (\bibinfo{year}{2017}).
\newblock


\bibitem[\protect\citeauthoryear{Harper and Konstan}{Harper and
  Konstan}{2016}]%
        {Harper2015The}
\bibfield{author}{\bibinfo{person}{F~Maxwell Harper} {and}
  \bibinfo{person}{Joseph~A Konstan}.} \bibinfo{year}{2016}\natexlab{}.
\newblock \showarticletitle{The movielens datasets: History and context}.
\newblock \bibinfo{journal}{{\em ACM Transactions on Interactive Intelligent
  Systems\/}} \bibinfo{volume}{5}, \bibinfo{number}{4} (\bibinfo{year}{2016}),
  \bibinfo{pages}{19}.
\newblock


\bibitem[\protect\citeauthoryear{He, Liao, Zhang, Nie, Hu, and Chua}{He
  et~al\mbox{.}}{2017}]%
        {he2017neural}
\bibfield{author}{\bibinfo{person}{Xiangnan He}, \bibinfo{person}{Lizi Liao},
  \bibinfo{person}{Hanwang Zhang}, \bibinfo{person}{Liqiang Nie},
  \bibinfo{person}{Xia Hu}, {and} \bibinfo{person}{Tat-Seng Chua}.}
  \bibinfo{year}{2017}\natexlab{}.
\newblock \showarticletitle{Neural collaborative filtering}. In
  \bibinfo{booktitle}{{\em Proceedings of the 26th International Conference on
  World Wide Web}}. \bibinfo{pages}{173--182}.
\newblock


\bibitem[\protect\citeauthoryear{Ioffe and Szegedy}{Ioffe and Szegedy}{2015}]%
        {Ioffe2015Batch}
\bibfield{author}{\bibinfo{person}{Sergey Ioffe} {and}
  \bibinfo{person}{Christian Szegedy}.} \bibinfo{year}{2015}\natexlab{}.
\newblock \showarticletitle{Batch normalization: Accelerating deep network
  training by reducing internal covariate shift}. In \bibinfo{booktitle}{{\em
  International conference on machine learning}}. \bibinfo{pages}{448--456}.
\newblock


\bibitem[\protect\citeauthoryear{Jain, Prabhu, and Varma}{Jain
  et~al\mbox{.}}{2016}]%
        {jain2016extreme}
\bibfield{author}{\bibinfo{person}{Himanshu Jain}, \bibinfo{person}{Yashoteja
  Prabhu}, {and} \bibinfo{person}{Manik Varma}.}
  \bibinfo{year}{2016}\natexlab{}.
\newblock \showarticletitle{Extreme multi-label loss functions for
  recommendation, tagging, ranking \& other missing label applications}. In
  \bibinfo{booktitle}{{\em Proceedings of the 22nd ACM SIGKDD International
  Conference on Knowledge Discovery and Data Mining}}. ACM,
  \bibinfo{pages}{935--944}.
\newblock


\bibitem[\protect\citeauthoryear{Jean, Cho, Memisevic, and Bengio}{Jean
  et~al\mbox{.}}{2014}]%
        {Jean2014On}
\bibfield{author}{\bibinfo{person}{S{\'e}bastien Jean},
  \bibinfo{person}{Kyunghyun Cho}, \bibinfo{person}{Roland Memisevic}, {and}
  \bibinfo{person}{Yoshua Bengio}.} \bibinfo{year}{2014}\natexlab{}.
\newblock \showarticletitle{On using very large target vocabulary for neural
  machine translation}.
\newblock \bibinfo{journal}{{\em arXiv preprint arXiv:1412.2007\/}}
  (\bibinfo{year}{2014}).
\newblock


\bibitem[\protect\citeauthoryear{Jin, Song, Li, Gai, Wang, and Zhang}{Jin
  et~al\mbox{.}}{2018}]%
        {jin2018real}
\bibfield{author}{\bibinfo{person}{Junqi Jin}, \bibinfo{person}{Chengru Song},
  \bibinfo{person}{Han Li}, \bibinfo{person}{Kun Gai}, \bibinfo{person}{Jun
  Wang}, {and} \bibinfo{person}{Weinan Zhang}.}
  \bibinfo{year}{2018}\natexlab{}.
\newblock \showarticletitle{Real-Time Bidding with Multi-Agent Reinforcement
  Learning in Display Advertising}.
\newblock \bibinfo{journal}{{\em arXiv preprint arXiv:1802.09756\/}}
  (\bibinfo{year}{2018}).
\newblock


\bibitem[\protect\citeauthoryear{Johnson, Douze, and J{\'e}gou}{Johnson
  et~al\mbox{.}}{2017}]%
        {JDH17}
\bibfield{author}{\bibinfo{person}{Jeff Johnson}, \bibinfo{person}{Matthijs
  Douze}, {and} \bibinfo{person}{Herv{\'e} J{\'e}gou}.}
  \bibinfo{year}{2017}\natexlab{}.
\newblock \showarticletitle{Billion-scale similarity search with GPUs}.
\newblock \bibinfo{journal}{{\em arXiv preprint arXiv:1702.08734\/}}
  (\bibinfo{year}{2017}).
\newblock


\bibitem[\protect\citeauthoryear{Koren, Bell, and Volinsky}{Koren
  et~al\mbox{.}}{2009}]%
        {Koren2009Matrix}
\bibfield{author}{\bibinfo{person}{Yehuda Koren}, \bibinfo{person}{Robert
  Bell}, {and} \bibinfo{person}{Chris Volinsky}.}
  \bibinfo{year}{2009}\natexlab{}.
\newblock \showarticletitle{Matrix Factorization Techniques for Recommender
  Systems}.
\newblock \bibinfo{journal}{{\em Computer\/}} \bibinfo{volume}{42},
  \bibinfo{number}{8} (\bibinfo{year}{2009}), \bibinfo{pages}{30--37}.
\newblock


\bibitem[\protect\citeauthoryear{Liang, Altosaar, Charlin, and Blei}{Liang
  et~al\mbox{.}}{2016}]%
        {Liang2016Factorization}
\bibfield{author}{\bibinfo{person}{Dawen Liang}, \bibinfo{person}{Jaan
  Altosaar}, \bibinfo{person}{Laurent Charlin}, {and} \bibinfo{person}{David~M.
  Blei}.} \bibinfo{year}{2016}\natexlab{}.
\newblock \showarticletitle{Factorization Meets the Item Embedding:Regularizing
  Matrix Factorization with Item Co-occurrence}. In \bibinfo{booktitle}{{\em
  ACM Conference on Recommender Systems}}. \bibinfo{pages}{59--66}.
\newblock


\bibitem[\protect\citeauthoryear{Lin}{Lin}{1999}]%
        {Lin1999WordNet}
\bibfield{author}{\bibinfo{person}{D. Lin}.} \bibinfo{year}{1999}\natexlab{}.
\newblock \showarticletitle{WordNet: An Electronic Lexical Database}.
\newblock \bibinfo{journal}{{\em Computational Linguistics\/}}
  \bibinfo{volume}{25}, \bibinfo{number}{2} (\bibinfo{year}{1999}),
  \bibinfo{pages}{292--296}.
\newblock


\bibitem[\protect\citeauthoryear{Linden, Smith, and York}{Linden
  et~al\mbox{.}}{2003}]%
        {linden2003amazon}
\bibfield{author}{\bibinfo{person}{Greg Linden}, \bibinfo{person}{Brent Smith},
  {and} \bibinfo{person}{Jeremy York}.} \bibinfo{year}{2003}\natexlab{}.
\newblock \showarticletitle{Amazon.com recommendations: Item-to-item
  collaborative filtering}.
\newblock \bibinfo{journal}{{\em IEEE Internet computing\/}}
  \bibinfo{volume}{7}, \bibinfo{number}{1} (\bibinfo{year}{2003}),
  \bibinfo{pages}{76--80}.
\newblock


\bibitem[\protect\citeauthoryear{Mikolov, Sutskever, Chen, Corrado, and
  Dean}{Mikolov et~al\mbox{.}}{2013}]%
        {Mikolov2013Distributed}
\bibfield{author}{\bibinfo{person}{Tomas Mikolov}, \bibinfo{person}{Ilya
  Sutskever}, \bibinfo{person}{Kai Chen}, \bibinfo{person}{Greg Corrado}, {and}
  \bibinfo{person}{Jeffrey Dean}.} \bibinfo{year}{2013}\natexlab{}.
\newblock \showarticletitle{Distributed representations of words and phrases
  and their compositionality}. In \bibinfo{booktitle}{{\em International
  Conference on Neural Information Processing Systems}}.
  \bibinfo{pages}{3111--3119}.
\newblock


\bibitem[\protect\citeauthoryear{Morin and Bengio}{Morin and Bengio}{2005}]%
        {Morin2005Hierarchical}
\bibfield{author}{\bibinfo{person}{Frederic Morin} {and}
  \bibinfo{person}{Yoshua Bengio}.} \bibinfo{year}{2005}\natexlab{}.
\newblock \showarticletitle{Hierarchical probabilistic neural network language
  model}.
\newblock \bibinfo{journal}{{\em Aistats\/}} (\bibinfo{year}{2005}).
\newblock


\bibitem[\protect\citeauthoryear{Ng, Jordan, and Weiss}{Ng
  et~al\mbox{.}}{2001}]%
        {Ng2001On}
\bibfield{author}{\bibinfo{person}{Andrew~Y. Ng}, \bibinfo{person}{Michael~I.
  Jordan}, {and} \bibinfo{person}{Yair Weiss}.}
  \bibinfo{year}{2001}\natexlab{}.
\newblock \showarticletitle{On spectral clustering: analysis and an algorithm}.
  In \bibinfo{booktitle}{{\em International Conference on Neural Information
  Processing Systems: Natural and Synthetic}}. \bibinfo{pages}{849--856}.
\newblock


\bibitem[\protect\citeauthoryear{Prabhu and Varma}{Prabhu and Varma}{2014}]%
        {prabhu2014fastxml}
\bibfield{author}{\bibinfo{person}{Yashoteja Prabhu} {and}
  \bibinfo{person}{Manik Varma}.} \bibinfo{year}{2014}\natexlab{}.
\newblock \showarticletitle{Fastxml: A fast, accurate and stable
  tree-classifier for extreme multi-label learning}. In
  \bibinfo{booktitle}{{\em Proceedings of the 20th ACM SIGKDD international
  conference on Knowledge discovery and data mining}}. ACM,
  \bibinfo{pages}{263--272}.
\newblock


\bibitem[\protect\citeauthoryear{Qu, Cai, Ren, Zhang, Yu, Wen, and Wang}{Qu
  et~al\mbox{.}}{2016}]%
        {Qu2016Product}
\bibfield{author}{\bibinfo{person}{Yanru Qu}, \bibinfo{person}{Han Cai},
  \bibinfo{person}{Kan Ren}, \bibinfo{person}{Weinan Zhang},
  \bibinfo{person}{Yong Yu}, \bibinfo{person}{Ying Wen}, {and}
  \bibinfo{person}{Jun Wang}.} \bibinfo{year}{2016}\natexlab{}.
\newblock \showarticletitle{Product-based neural networks for user response
  prediction}. In \bibinfo{booktitle}{{\em IEEE 16th International Conference
  on Data Mining}}. IEEE, \bibinfo{pages}{1149--1154}.
\newblock


\bibitem[\protect\citeauthoryear{Rendle}{Rendle}{2010}]%
        {Rendle2010Factorization}
\bibfield{author}{\bibinfo{person}{Steffen Rendle}.}
  \bibinfo{year}{2010}\natexlab{}.
\newblock \showarticletitle{Factorization Machines}. In
  \bibinfo{booktitle}{{\em IEEE International Conference on Data Mining}}.
  \bibinfo{pages}{995--1000}.
\newblock


\bibitem[\protect\citeauthoryear{Rendle, Freudenthaler, Gantner, and
  Schmidt-Thieme}{Rendle et~al\mbox{.}}{2009}]%
        {rendle2009bpr}
\bibfield{author}{\bibinfo{person}{Steffen Rendle}, \bibinfo{person}{Christoph
  Freudenthaler}, \bibinfo{person}{Zeno Gantner}, {and} \bibinfo{person}{Lars
  Schmidt-Thieme}.} \bibinfo{year}{2009}\natexlab{}.
\newblock \showarticletitle{BPR: Bayesian personalized ranking from implicit
  feedback}. In \bibinfo{booktitle}{{\em Proceedings of the 25th conference on
  uncertainty in artificial intelligence}}. AUAI Press,
  \bibinfo{pages}{452--461}.
\newblock


\bibitem[\protect\citeauthoryear{Salakhutdinov and Mnih}{Salakhutdinov and
  Mnih}{2007}]%
        {Salakhutdinov2007Probabilistic}
\bibfield{author}{\bibinfo{person}{Ruslan Salakhutdinov} {and}
  \bibinfo{person}{Andriy Mnih}.} \bibinfo{year}{2007}\natexlab{}.
\newblock \showarticletitle{Probabilistic Matrix Factorization}. In
  \bibinfo{booktitle}{{\em International Conference on Neural Information
  Processing Systems}}. \bibinfo{pages}{1257--1264}.
\newblock


\bibitem[\protect\citeauthoryear{Sarwar, Karypis, Konstan, and Riedl}{Sarwar
  et~al\mbox{.}}{2001}]%
        {Sarwar2001Item}
\bibfield{author}{\bibinfo{person}{Badrul Sarwar}, \bibinfo{person}{George
  Karypis}, \bibinfo{person}{Joseph Konstan}, {and} \bibinfo{person}{John
  Riedl}.} \bibinfo{year}{2001}\natexlab{}.
\newblock \showarticletitle{Item-based collaborative filtering recommendation
  algorithms}. In \bibinfo{booktitle}{{\em International Conference on World
  Wide Web}}. \bibinfo{pages}{285--295}.
\newblock


\bibitem[\protect\citeauthoryear{Weston, Makadia, and Yee}{Weston
  et~al\mbox{.}}{2013}]%
        {weston2013label}
\bibfield{author}{\bibinfo{person}{J. Weston}, \bibinfo{person}{A. Makadia},
  {and} \bibinfo{person}{H. Yee}.} \bibinfo{year}{2013}\natexlab{}.
\newblock \showarticletitle{Label partitioning for sublinear ranking}. In
  \bibinfo{booktitle}{{\em International Conference on Machine Learning}}.
  \bibinfo{pages}{181--189}.
\newblock


\bibitem[\protect\citeauthoryear{Xu, Wang, Chen, and Li}{Xu
  et~al\mbox{.}}{2015}]%
        {Xu2015Empirical}
\bibfield{author}{\bibinfo{person}{Bing Xu}, \bibinfo{person}{Naiyan Wang},
  \bibinfo{person}{Tianqi Chen}, {and} \bibinfo{person}{Mu Li}.}
  \bibinfo{year}{2015}\natexlab{}.
\newblock \showarticletitle{Empirical evaluation of rectified activations in
  convolutional network}.
\newblock \bibinfo{journal}{{\em arXiv:1505.00853\/}} (\bibinfo{year}{2015}).
\newblock


\bibitem[\protect\citeauthoryear{Zhou, Song, Zhu, Ma, Yan, Dai, Zhu, Jin, Li,
  and Gai}{Zhou et~al\mbox{.}}{2018}]%
        {Zhou2017Deep}
\bibfield{author}{\bibinfo{person}{Guorui Zhou}, \bibinfo{person}{Chengru
  Song}, \bibinfo{person}{Xiaoqiang Zhu}, \bibinfo{person}{Xiao Ma},
  \bibinfo{person}{Yanghui Yan}, \bibinfo{person}{Xingya Dai},
  \bibinfo{person}{Han Zhu}, \bibinfo{person}{Junqi Jin}, \bibinfo{person}{Han
  Li}, {and} \bibinfo{person}{Kun Gai}.} \bibinfo{year}{2018}\natexlab{}.
\newblock \showarticletitle{Deep interest network for click-through rate
  prediction}. In \bibinfo{booktitle}{{\em Proceedings of the 24th ACM SIGKDD
  Conference}}. \bibinfo{publisher}{ACM}.
\newblock


\bibitem[\protect\citeauthoryear{Zhu, Jin, Tan, Pan, Zeng, Li, and Gai}{Zhu
  et~al\mbox{.}}{2017}]%
        {OCPC}
\bibfield{author}{\bibinfo{person}{Han Zhu}, \bibinfo{person}{Junqi Jin},
  \bibinfo{person}{Chang Tan}, \bibinfo{person}{Fei Pan},
  \bibinfo{person}{Yifan Zeng}, \bibinfo{person}{Han Li}, {and}
  \bibinfo{person}{Kun Gai}.} \bibinfo{year}{2017}\natexlab{}.
\newblock \showarticletitle{Optimized Cost Per Click in Taobao Display
  Advertising}. In \bibinfo{booktitle}{{\em Proceedings of the 23rd ACM SIGKDD
  Conference}}. \bibinfo{publisher}{ACM}, \bibinfo{pages}{2191--2200}.
\newblock


\end{thebibliography}

\end{document}